\documentclass{article}

\PassOptionsToPackage{numbers, sort&compress}{natbib}

\usepackage[preprint]{nips_2018}

\usepackage[utf8]{inputenc} 
\usepackage[T1]{fontenc}    
\usepackage{hyperref}       
\usepackage{url}            
\usepackage{booktabs}       
\usepackage{amsfonts}       
\usepackage{nicefrac}       
\usepackage{microtype}      
\usepackage{graphicx}
\usepackage{subcaption}
\usepackage{amsmath}
\usepackage{rotating}
\usepackage{floatpag}

\title{Dynamic learning rate using Mutual Information}

\author{Shrihari Vasudevan\\
		    IBM Research - India\\
			  \texttt{shrivasu@in.ibm.com}}

\begin{document}

\maketitle

\begin{abstract}
	This paper demonstrates dynamic hyper-parameter setting, for deep neural network training, using Mutual Information (MI). The specific hyper-parameter studied in this paper is the learning rate. MI between the output layer and true outcomes is used to dynamically set the learning rate of the network through the training cycle; the idea is also extended to layer-wise setting of learning rate. Two approaches are demonstrated - tracking relative change in mutual information and, additionally tracking its value relative to a reference measure. The paper does not attempt to recommend a specific learning rate policy. Experiments demonstrate that mutual information may be effectively used to dynamically set learning rate and achieve competitive to better outcomes in competitive to better time.
\end{abstract}

\section{Introduction}

Hyper-parameter selection in deep neural networks is mostly done by experimentation for different data-sets and models. A key example of one such hyper-parameter that is the subject of this paper is the learning rate. High learning rates, particularly in early training stages, can result in instabilities and fluctuations in the parameter search process. Established procedures to set learning rate to a low value at the beginning and then gradually ``warm-up''  to the desired learning rate have been used effectively \cite{He2016, Goyal2017}. These approaches require the a-priori definition of a policy or schedule and the learning rate changes according to that fixed policy. The fixed policy may not be suited for different data-sets or model architectures which may be very different in complexity. The same policy may also not suffice in case the compute resources available, or other factors, necessitate changes in other training hyper-parameters (e.g. batch size) during training. Dynamically setting the learning rate through the training cycle is one way of handling such issues; this paper studies the feasibility of Mutual Information (MI) \cite{Cover2006} as a metric to realize this objective.

\section{Related work}

Adaptive learning rate schedules based on gradients have been proposed in various gradient-descent based optimization algorithms used for training deep neural networks \cite{ruder2016}. These include the likes of AdaGrad, AdaDelta, RMSprop, Adam and some more recent algorithms. These set learning rates at the level of individual parameters by considering the frequency or magnitude of updates; slow or infrequent updates characterized by smaller past gradients get more importance than fast or frequent updates characterized by larger past gradients. Depending on the data-set and model complexity, careful initial selection of the learning rate may be required.

Recent works of Tishby et al \cite{Tishby2015, Schwartz-Ziv2017} have attempted to explain deep learning through the Information Bottleneck (IB) \cite{Tishby1999} basis. In particular, the recent paper \cite{Schwartz-Ziv2017} made strong and wide-ranging claims on aspects relating to phases in deep learning, causal relationship between compression and generalization and the basis for compression in deep learning. Some of these claims were subsequently countered in \cite{Saxe2018}, while acknowledging the potential of the more general MI and IB concepts. The current paper builds on this body of literature.

An important and related problem is that of estimation of MI. The problem has been widely studied and several estimation methods exist \cite{walters-williams2009}. One such method is that of nearest neighbor approaches; these have been shown \cite{doquire2012,khan2007} to be effective with high dimensional data and at large sample sizes. A widely cited example of this class of algorithms is the Kraskov-St\"ogbauer-Grassberger (KSG) estimator \cite{kraskov2004}. This approach is used for MI estimation in this paper; other algorithms can also be used. Recent developments in the area include \cite{Kolchinsky2017} that estimates MI using pairwise distances between Gaussian mixture components and \cite{belghazi2018} that estimates MI through gradient descent over neural networks. Their properties and suitability to the context of this paper will be studied in future.

This work seeks to understand the operational utility of MI as a metric for deep learning and specifically, for dynamic hyper-parameter (in this paper, the learning rate) setting. For this paper, the essential core of a typical deep neural network training pipeline remains unchanged i.e. the use of mini-batch Stochastic Gradient Descent (SGD) optimization to optimize (minimize) the training cost (e.g. mis-classification error) is maintained but the learning rate is set adaptively considering the MI of hidden layer activations with the true output. The paper effectively demonstrates that an information driven ``warm-up'' and subsequent ``cool-down'' of learning rate can produce competitive outcomes on standard data-sets; it does not attempt to prescribe a specific learning rate policy. 

\section{Approach}

In (deep) neural network models, MI lends a layer specific measure that may be utilized in multiple ways. It may potentially be used as a metric for parameter optimization through standard optimization methods used for the purpose \cite{Tishby2015}; works such as \cite{Shamir2010} suggested MI as providing an upper bound for prediction error. It may serve as a basis for dynamic tuning of network-level hyper-parameters. Further, as a layer-specific measure, it may be utilized for layer-wise dynamic tuning of hyper-parameters. Based on the information metric per epoch, interventions in hyper-parameters may be used to steer the learning process towards efficient and effective deep learning. 

\begin{figure}
	\begin{subfigure}[t]{0.49\textwidth}
		\includegraphics[width=\textwidth]{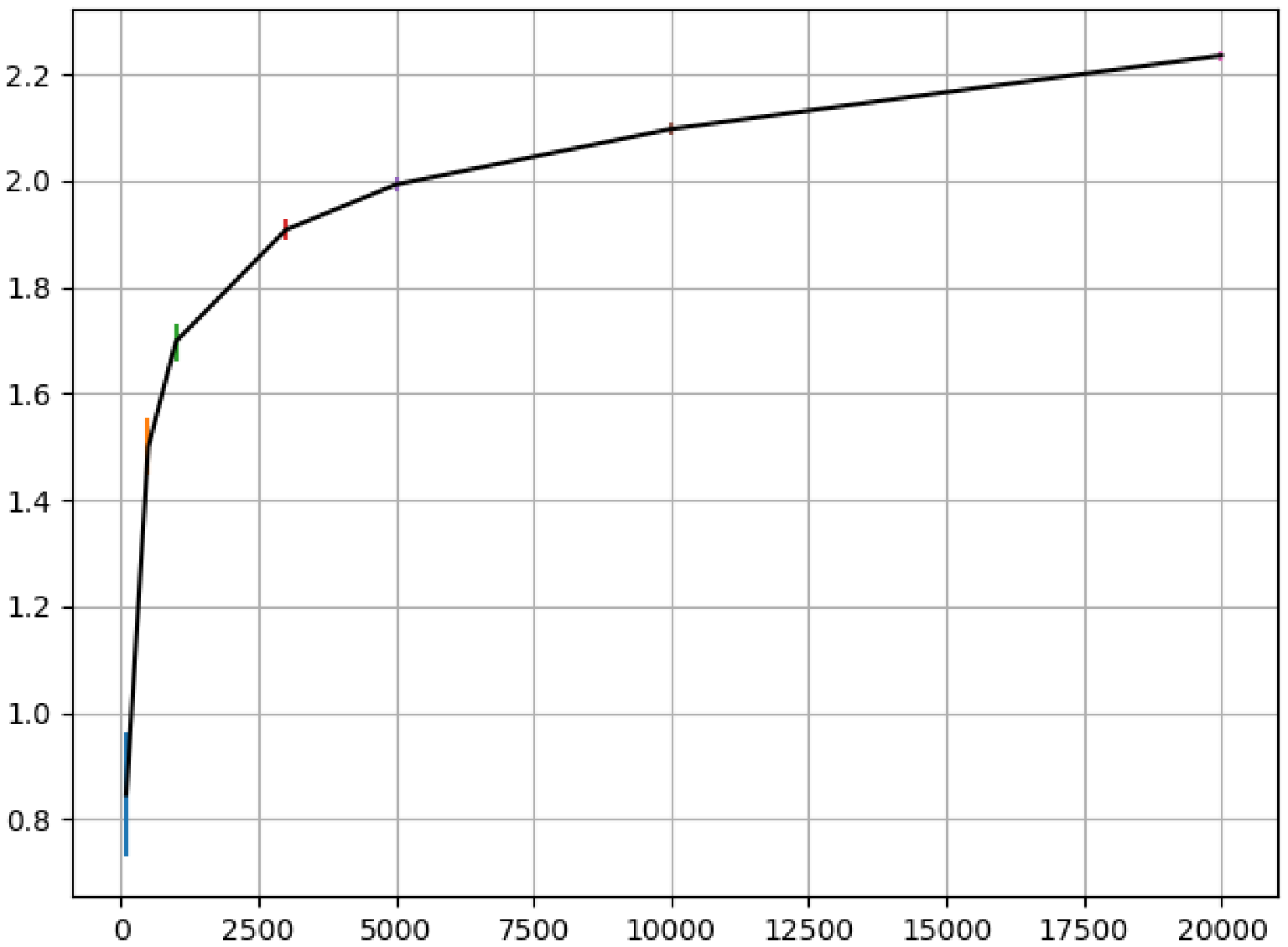}
	\end{subfigure}
	\begin{subfigure}[t]{0.49\textwidth}
		\includegraphics[width=\textwidth]{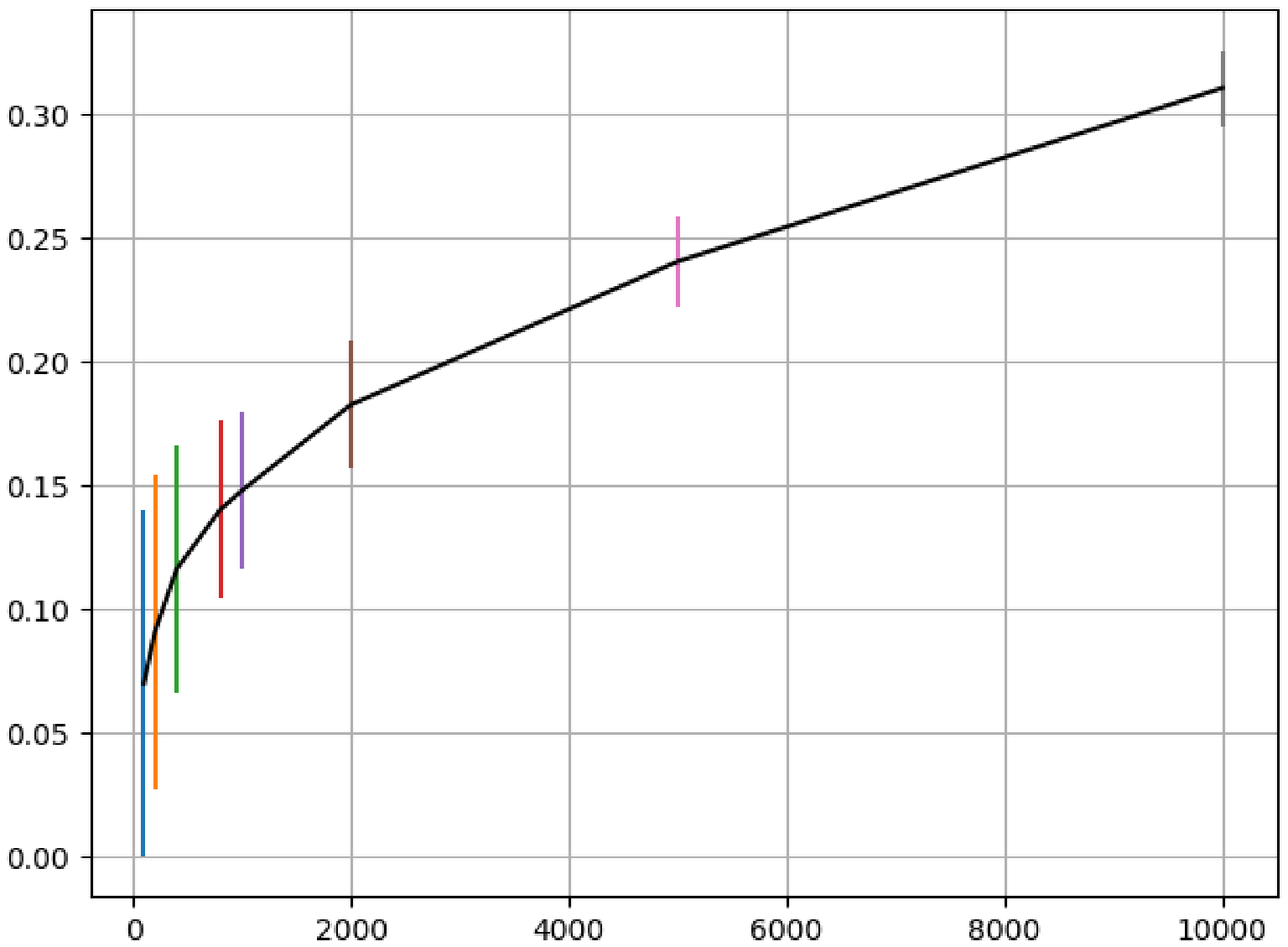}
	\end{subfigure}
	\caption{MI (of input and output training data) vs sample size for MNIST (left) and CIFAR-10 (right) as computed using the KSG estimator. The figures show estimated mean and standard deviation (error bar) for each sample size tested. Experiments in this paper use a sample size of 1000.}
	\label{fig:mi_vs_samplesize}
\end{figure}

Computing MI is generally computationally expensive; performing MI computation after each epoch in deep neural net training can prove to be infeasible. This paper relies on two ideas to effectively use MI in training with large data sets - (1) use a randomly selected subset of data for MI computation - plotting the MI vs sample size curve for different data sets (see Fig \ref{fig:mi_vs_samplesize}) enables informed selection of an appropriate subset sample-size for per-epoch MI computation and (2) this approximate MI value may suffice if the relative measures can be utilized for the problem being addressed.

Given input $X$, output $Y$ and hidden layer activations $H_i$, MI between input and output, $I(X;Y)$, is denoted as $IXY$. MI between hidden layer activations and output, $I(H_i;Y)$, is denoted by as $IHY$; specifically, MI between last (output) layer activations and output is denoted as $IHYLL$. Finally, MI between hidden layer activations and input, $I(H_i;X)$, is denoted by as $IHX$.

In the context of neural networks, the Data Processing Inequality (DPI) \cite{Cover2006} effectively provides an upper bound to the information that each layer (including output) of a neural network can capture. It suggests that successive layers $H_i$ operating on the input data $X$ cannot increase its information content relative to the output data i.e. $IHY \leq IXY$; as a specific case, $IHYLL \leq IXY$. 

With $IXY$ computed from input data $X$ and output data $Y$, this inequality may hold true for dense neural networks but will not hold true for convolution neural networks (CNNs). The use of multiple (say $n$) convolution filters in CNNs is akin to treating $n$ image inputs tiled together with the respective convolution filter weights being mapped to the weights of a much larger dense neural network. Thus, a reasonable estimate of the upper bound $IXY$ may be obtained by repeating or tiling $X$ and $Y$, $n$ times, and then computing the estimate $IXY$. 

The work \cite{Schwartz-Ziv2017} suggests $IHYLL/IXY$ as a ratio of the amount of information captured by the model. The KSG estimator \cite{kraskov2004} incorporates a small amount of noise (a jitter in the order of $10^{-10}$) in MI computation to overcome degeneracies in data. Saxe et al \cite{Saxe2018} observe that DPI will not hold when noise is added for the purposes of measuring MI. As a consequence of the DPI not being valid, $IXY$ may not be an upper bound and $IHY > IXY$ is a possible outcome. This paper proposes that the upper bound $IXY$ may be used as a ``soft'' criterion for dynamic learning rate setting using MI. Specifically, this paper proposes to increase the learning rate towards achieving the soft upper bound of $IXY$ and using the DPI violation condition ($IHY > IXY$) as a signal to reduce learning rate.

While adaptive learning rates schedules have been developed using gradients \cite{ruder2016} and can, in principle, be developed using other measures such as validation accuracy, the use of MI as a criterion for dynamic tuning of hyper-parameters is motivated by this measure being able to capture both linear and nonlinear dependence between the quantities of interest (in our case, hidden layer activations and the true output) and offer a layer-wise measure of optimality ($IHY$) with respect to a reference measure ($IXY$). This paper uses standard deep neural net training design choices (e.g. mini-batch SGD to minimize misclassification error) while setting the learning rate to maximize information with respect to true outcomes i.e. maximize $IHY$. 

Given the soft upper bound of $IXY$, training the neural network increases the information of the last layer with respect to the true outcomes $IHYLL$ until it finally saturates to a maximum. This trend is also observed for previous layers though the change over the training cycle may be less dramatic and the $IHY$ value is typically lower. This observation serves as the basis for dynamically setting the learning rate (LR) in this paper. Two approaches are explored - 
\begin{enumerate}
	\item Tracking the change in $IHYLL$, denoted by $\delta IHYLL$, relative to its value
	\begin{itemize}
		\item This basically uses behavior that when the information measure saturates, change in the measure diminishes. The relative change measure $\frac{\delta IHYLL}{IHYLL} < \epsilon$ as $IHYLL$ saturates; $\epsilon$ is a small number. 
	\end{itemize}
	\item Tracking $IHYLL$ and $\delta IHYLL$ relative to $IXY$
		\begin{itemize}
			\item LR increases and decreases are set in terms of the relative measure $\frac{IHYLL}{IXY}$ so as to maximize $IHYLL$ relative to $IXY$; this measure coupled with the relative change $\frac{\delta IHYLL}{IXY}$ between epochs are used to decide on increases or decreases in LR.
		\end{itemize}
\end{enumerate}
Both approaches require the specification of a minimum and maximum LR (upper and lower bounds) and begin from the minimum value; experiments in this paper set these bounds around the desired LR. The first approach increases LR while the relative change in $IHYLL$ is significant (to a threshold, $\epsilon$); thereafter LR decreases. The second approach tracks both the value $IHYLL$ and the change in $IHYLL$ between epochs, relative to $IXY$. LR increases while significant changes in $IHYLL$ occur and $IHYLL \leq IXY$ and thereafter, it decreases. In both cases, LR changes incrementally to enable gradual changes. LR increases may be performed at the same rate as decreases or may be dampened to control the max LR reached. LR policies used in the experiments are provided in the appendix. Note that this paper does not propose a specific learning rate policy; it focuses on demonstrating that MI can be used to dynamically set the LR to achieve competitive outcomes.

\section{Experiments}

\begin{figure}
	\includegraphics[width=\textwidth]{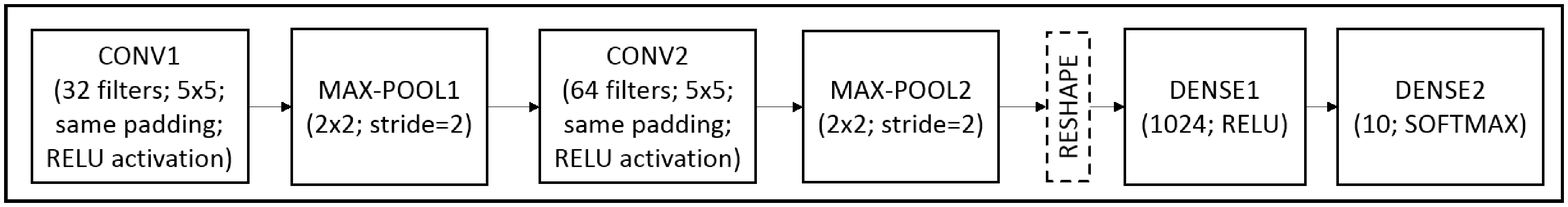}
	\caption{Model used for experiments on MNIST data-set}
	\label{fig:model}
\end{figure}

Two standard data-sets were used to demonstrate dynamic LR setting using MI - MNIST \cite{Lecun1998} and CIFAR-10 \cite{Krizhevsky2009}. The emphasis (model and design choices) of these experiments was \emph{not} maximizing accuracy for the data-set but to understand relative performance of dynamic LR setting (using MI) in comparison to alternatives. The model used for the experiments with MNIST data is shown in Figure \ref{fig:model}; these experiments used standard off-the-shelf mini-batch SGD to minimize mis-classification error (categorical cross entropy). For CIFAR-10 experiments, the model used was based on \cite{springenberg2015}. The specific model implementation used dropout (50\%) only after max-pooling layers, no L2 regularization for weights, a fixed momentum value of 0.9 and Nesterov acceleration. With a continuously decaying LR beginning at 0.01, it resulted in sufficiently close maximum test accuracies of 88.21\% and 91.23\%, with and without data augmentation respectively, to the respective outcomes (90.92\% and 92.75\%) over 350 epochs in the referenced paper. Experiments of this paper used the implementation without any data augmentation to enable a fair comparison between methods.

\begin{figure}
	\centering
	\captionsetup{width=1.4\linewidth}
	\makebox[\textwidth]{\includegraphics[width=1.4\textwidth]{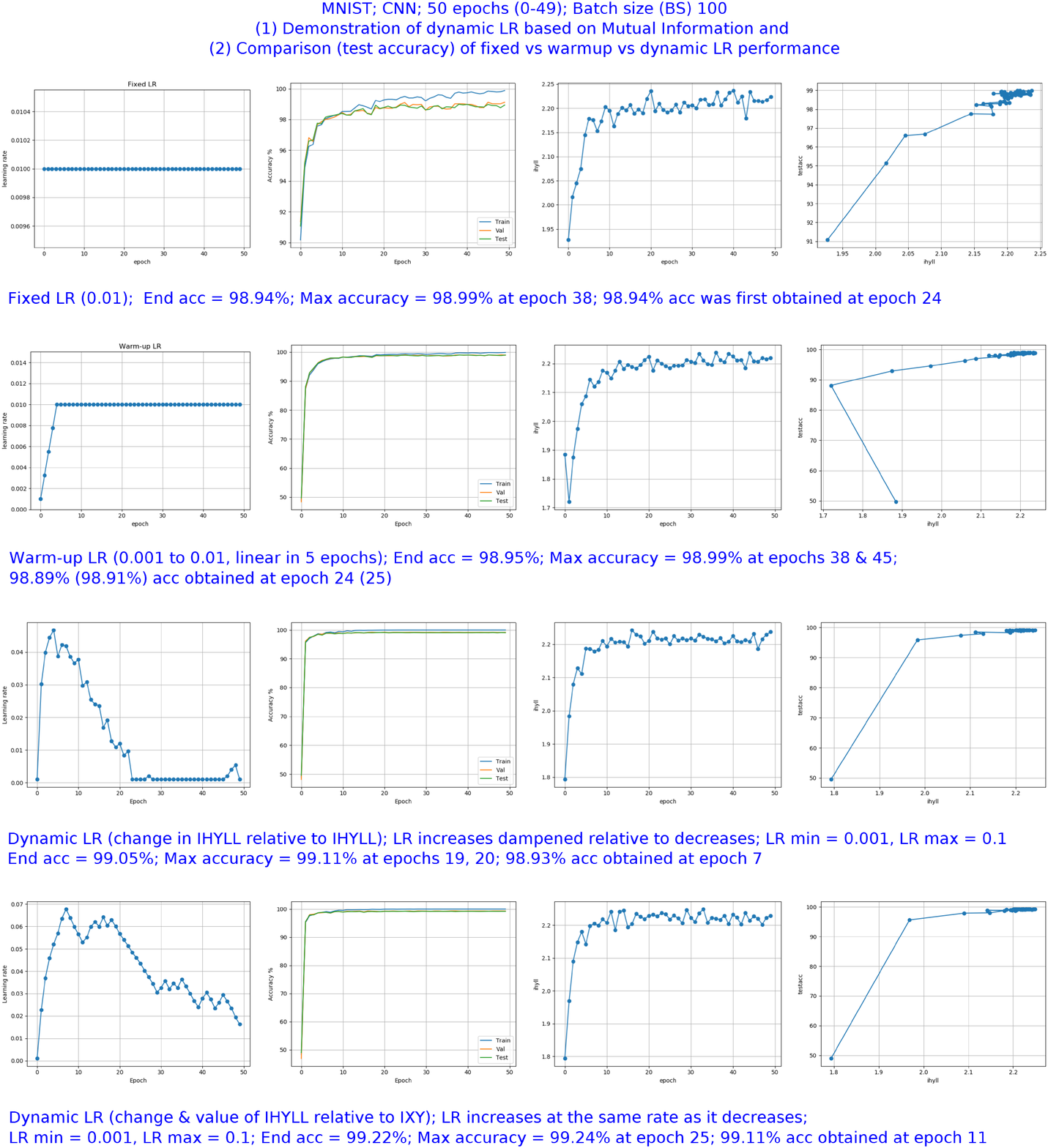}}
	\caption{MNIST - comparison of fixed, warm-up and dynamic LR methods. Each row shows four plots - LR vs epoch, accuracy vs epoch, IHYLL vs epoch and test-accuracy vs IHYLL. Dynamic LR is able to start low and gradually explore a larger range of learning rates as required by the data/model and then cools down to produce better outcomes in better time. Tracking both change and value of $IHYLL$ relative to $IXY$ performed better than tracking the relative change in $IHYLL$ alone.}
	\label{fig:mnist}
\end{figure}

\begin{figure}
	\centering
	\captionsetup{width=1.4\linewidth}
        \floatpagestyle{empty}
	\vspace{-2cm}
	\makebox[\textwidth]{\includegraphics[width=1.32\textwidth]{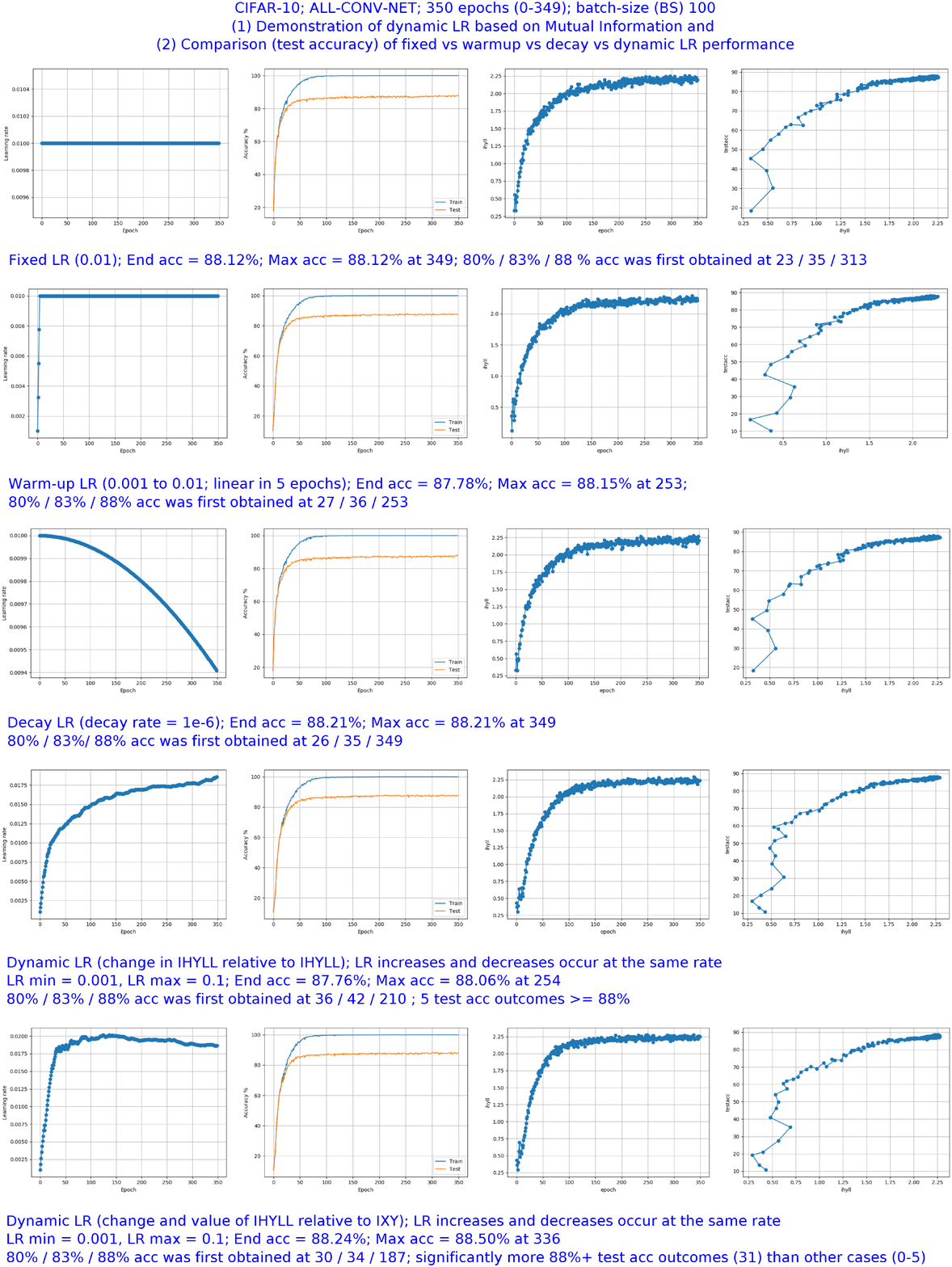}}
	\caption{CIFAR-10 - comparison of fixed, warm-up, decay and dynamic LR methods using a model based on \cite{springenberg2015}. Each row shows four plots - LR vs epoch, accuracy vs epoch, IHYLL vs epoch and test-accuracy vs IHYLL. Dynamic LR is able to start low and gradually explore a larger range of learning rates as required by the data/model to produce better outcomes in better time. Tracking both change and value of $IHYLL$ relative to $IXY$ performed better than tracking the relative change in $IHYLL$ alone.}
	\label{fig:cifar10}
\end{figure}

Figures \ref{fig:mnist} and \ref{fig:cifar10} show outcomes of various LR selection methods applied to the MNIST and CIFAR-10 data-sets respectively; this experiment compared the use of a fixed (desired) LR, a warm-up beginning from a lower LR and increasing to the desired LR and the dynamic LR methods tracking either the change in $IHYLL$ relative to its value or tracking both the value and the change in $IHYLL$ relative to the reference measure of $IXY$. The CIFAR-10 experiment also adds a decaying LR policy into the comparison. The following trends were observed - 
\begin{itemize}
	\item Training deep neural networks results in increasing $IHYLL$. This is and intuitive and expected outcome of successful training.
	\item There is a non-linear and increasing relationship between test accuracy and $IHYLL$. Increasing $IHYLL$ results in higher test accuracy.
	\item Achieving maximum $IHYLL$ does not necessarily guarantee (from existing plots) maximum test accuracy but definitely gives a very competitive test accuracy. Per the Information Bottleneck approach, the same value of $IHY$ may be associated with different $IHX$ and it is likely that a more compressed (lower $IHX$) model generalizes better. 
	\item It appears that an effective warm-up should ideally result in significant increase in $IHYLL$ towards $IXY$. The warm-up policy (linear increase to desired learning rate in 5 epochs) seems effective for MNIST and insufficient for CIFAR-10.
        \item Experiments reported in this paper and other attempts suggest that the MNIST was able to produce good outcomes with an aggressive (relative to CIFAR-10) warm-up and cool-down LR policy; CIFAR-10 on the other hand required the use of a slow warm-up and lower LR values over-all. 
	\item It is clear that mutual information of hidden layer activations with respect to output may be useful for dynamically setting learning rate through the training process to obtain competitive to better test accuracies in competitive to better time. A policy involving both change in $IHYLL$ and its value relative to $IXY$ produces better outcomes than tracking the former alone. In both MNIST (Figure \ref{fig:mnist}) and CIFAR-10 (Figure \ref{fig:cifar10}), dynamic LR using both the change and the value of $IHYLL$ resulted in top accuracy levels being achieved in roughly half the number of epochs as compared to the corresponding fixed LR policy.
        \item A dynamic LR policy based on MI makes training easier in the sense that it moves this hyper-parameter selection problem one level up; the problems of specifying a single optimal LR for the entire training cycle or specifying an optimal warm-up policy to a "good" LR are overcome by automatically adjusting the LR every epoch, between bounds, to effectively realize an information driven warm-up and cool-down of the LR. Starting from a low value of learning rate results in a stable search process and outcomes. Competitive outcomes are achieved by exploring a larger space of learning rates than the fixed and warm-up strategies which are both essentially fixed learning rate policies; it is also also possible to achieve competitive outcomes using a smaller learning rate for a longer length of time.
\end{itemize}

\begin{sidewaysfigure}
	\includegraphics[width=\textwidth]{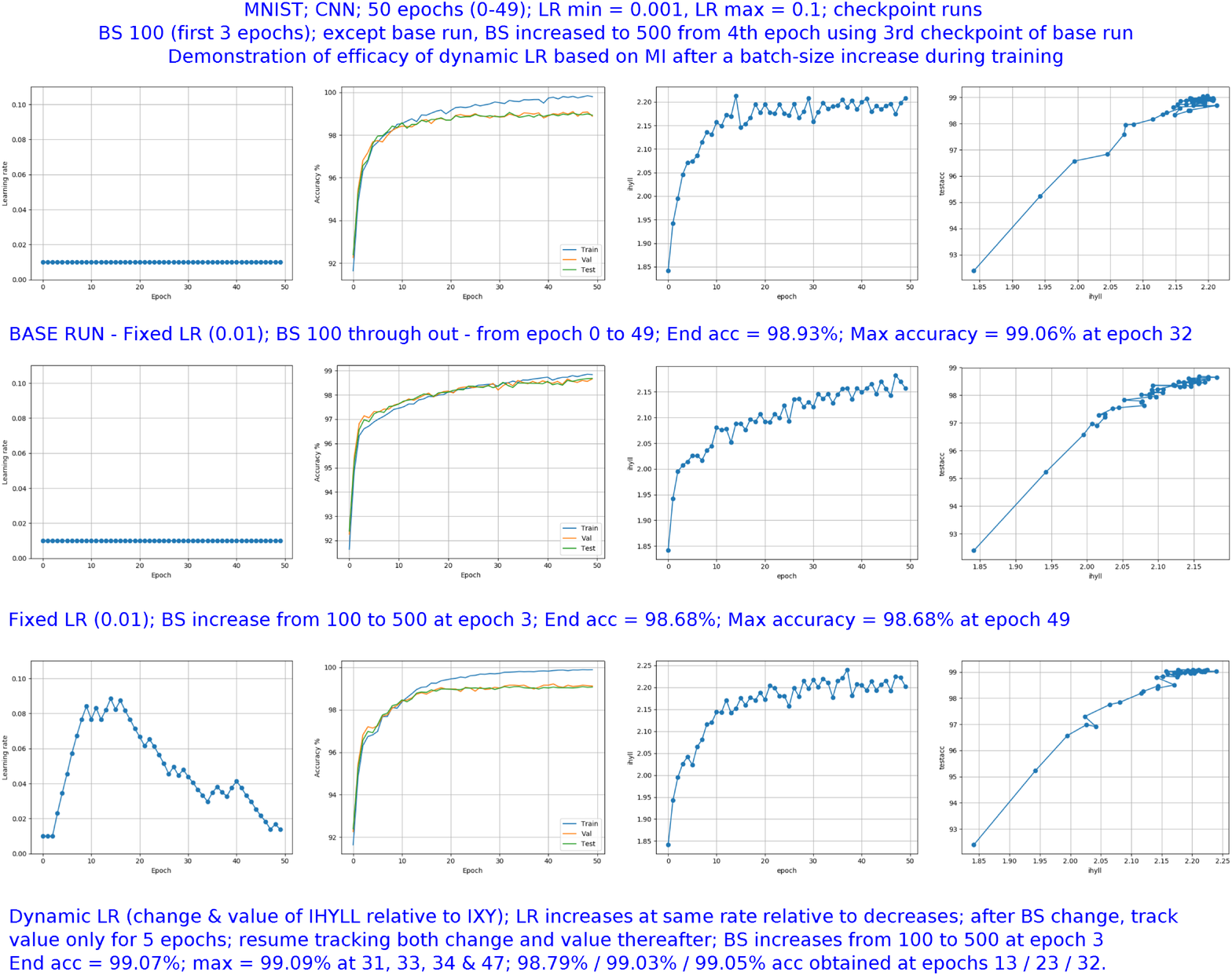}
	\caption{MNIST - dynamic LR using mutual information produces competitive to better outcomes when another hyper-parameter (e.g. batch size BS) changes during training. The base run does not perform a BS change; the other two runs are started from the 3rd checkpoint of this run. Each row in the figure shows LR vs epoch, accuracy vs epoch, IHYLL vs epoch and test-accuracy vs IHYLL.}
	\label{fig:mnist_bs_increase}
\end{sidewaysfigure}

\begin{sidewaysfigure}
	\includegraphics[width=\textwidth]{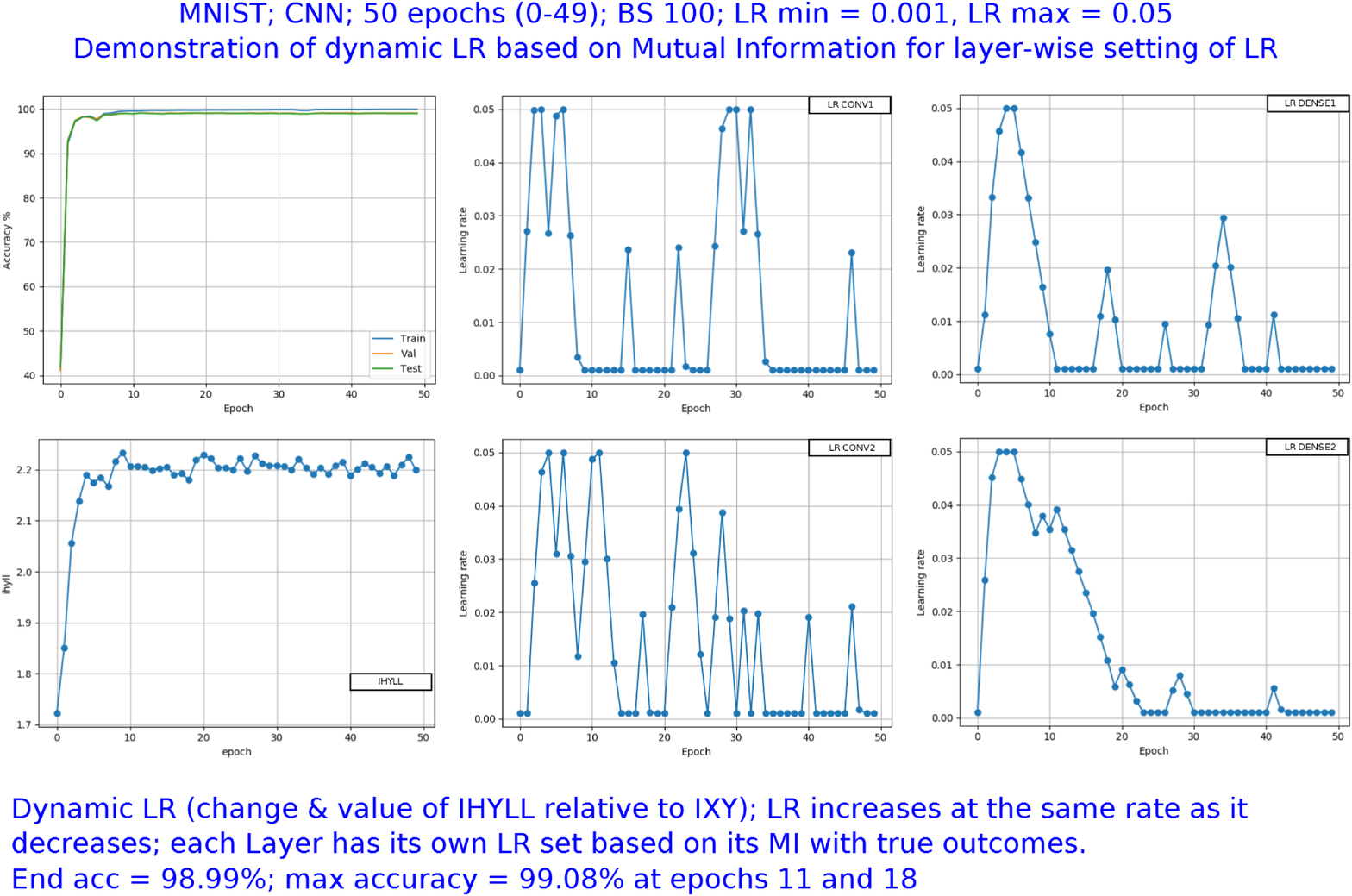}
	\caption{MNIST - layer-wise dynamic LR using MI produces competitive outcomes in competitive time as compared to single LR training. Left column shows accuracy vs epoch and IHYLL vs epoch, middle column shows the LR vs epoch for the convolution layers and right column shows the LR vs epoch for the dense layers.}
	\label{fig:mnist_layerwise_dynamicLR}
\end{sidewaysfigure}

Figure \ref{fig:mnist_bs_increase} shows an application experiment where the dynamic LR concept may be useful. During a training run, if the situation (e.g. availability of compute resources or simply, a training design choice) requires an increase in one hyper-parameter such as the batch-size (BS), the LR would have to be suitably increased or a drop in accuracy may occur. There are guidelines on managing the LR in such scenarios. This experiment however, demonstrates that dynamic LR based on MI can be effectively used to automatically adjust LR in such scenarios. The LR policy used here tracked only the value of $IHYLL$ relative to $IXY$ for a few epochs before resuming the tracking of both change and value; this was done to enable the growth of both LR and $IHYLL$ as a consequence of the increased batch size before resuming the tracking of both its change and value. Note the higher LR reached in this experiment as compared to the fixed batch size run in Figure \ref{fig:mnist}. Competitive to better outcomes were achieved in competitive to better time. The availability of a reference measure (a soft upper bound) $IXY$ enables MI to be particularly suited to handle such scenarios, as compared to other more readily available measures.

The use of MI of the last layer alone $IHYLL$ provides a network-level intervention i.e. dynamic LR setting for all layers. The key property that mutual information affords is a layer-wise measure of optimality. The methods demonstrated thus far were extended to a layer-wise intervention i.e. dynamic LR setting of individual layers; this is demonstrated in Figure \ref{fig:mnist_layerwise_dynamicLR} where each layer's MI with the true outcomes $IHY$ relative to the reference measure $IXY$ enables the setting of a layer-specific LR. Competitive outcomes were obtained in competitive time, compared to the outcomes of Figure \ref{fig:mnist}.

\section{Conclusion}

This paper demonstrated that using Mutual Information (MI) to dynamically set learning rate through the training cycle, in the context of deep neural networks, is both feasible and produces competitive to better outcomes in competitive to better time. The paper also demonstrated the application of this idea to automatically respond to changes in other hyper-parameters such as the batch-size and the extension of the idea to a layer-wise dynamic LR tuning through the training cycle. MI lends a layer-wise measure of optimality with respect to a reference value that can be leveraged to effectively steer deep neural network training to competitive/better outcomes. 

\section*{Appendix}

The following policies were used in experiments of this paper. Note that the paper does not attempt to prescribe a specific learning rate policy.

Dynamic LR policy based on change in $IHYLL$
\begin{align}
\delta_t &= |(IHYLL_{t-1}-IHYLL_{t-2})|/IHYLL_{t-1} \nonumber\\
LR_t &= \begin{cases}
min(LR_{max}, LR_{t-1} + \gamma_1*\delta_t) \;\;\;(\delta_t > \epsilon) \\
max(LR_{min}, LR_{t-1} - \gamma_2*\delta_t)	\;\;\;(\delta_t \leq \epsilon)
\end{cases} \nonumber
\end{align}		

$LR_{min}$ and $LR_{max}$ are selected by the user for each data set. The LR for the current epoch, $LR_t$, is decided based on that of the previous epoch, $LR_{t-1}$, and the relative change in $IHYLL$ with respect to its value. $\epsilon$ is a small number e.g. 0.01. The $\gamma$ parameters allow dampening LR increases relative to decreases, if required; for e.g., $\gamma_1$ = 0.1 and $\gamma_2$ = 1 was used for MNIST and $\gamma_1$ = 0.003 and $\gamma_2$ = 0.003, for CIFAR-10.

Dynamic LR policy based on change and value of $IHYLL$ relative to $IXY$
\begin{align}
d1_t &= 1-(IHYLL_{t-1}/IXY) \nonumber\\
d2_t &= |(IHYLL_{t-1}-IHYLL_{t-2})|/IHYLL_{t-1} \nonumber\\
LR_t &= 
	\begin{cases}
		min(LR_{max}, LR_{t-1} + \gamma_1*d1_t)    \;\;(d1_t > 0 \;\&\; d2_t > \epsilon)\\	
		max(LR_{min}, LR_{t-1} - \gamma_2*d1_t)	\;\;(d1_t > 0 \;\&\; d2_t \leq \epsilon)\\
		max(LR_{min}, LR_{t-1} + \gamma_3*d1_t)	\;\;(d1_t \leq 0 \;\&\; d2_t > \epsilon)\\
		max(LR_{min}, LR_{t-1} + \gamma_3*d1_t)	\;\;(d1_t \leq 0 \;\&\; d2_t \leq \epsilon)
	\end{cases} \nonumber
\end{align}		

The terms are defined as before. There are effectively two LR regimes governed by $d1_t$ being $>0$ or $\leq0$; in the former the LR may increase or decrease depending on whether $d2_t$ saturates ($\leq\epsilon$) or not; the latter case involves LR reductions only. For both MNIST and CIFAR-10, LR increases occurred at the same rate as decreases - for MNIST, $\gamma_1=\gamma_2=0.1$ and for CIFAR-10, $\gamma_1=\gamma_2=0.001$; for both MNIST and CIFAR-10, $\gamma_3$ was set to 0.1.

\small
\bibliographystyle{unsrtnat}
\bibliography{ref}

\end{document}